\documentclass[letterpaper]{article} 
\usepackage{aaai2026}  
\usepackage{times}  
\usepackage{helvet}  
\usepackage{courier}  
\usepackage[hyphens]{url}  
\usepackage{graphicx} 
\usepackage{natbib}  
\usepackage{caption} 
\usepackage{tcolorbox}
\usepackage{xcolor}

\usepackage{algorithm}
\usepackage{algorithmic}
\usepackage{amsmath}
\usepackage{amssymb}
\usepackage{tabularx}
\usepackage{booktabs}

\usepackage{newfloat}
\usepackage{listings}
\DeclareCaptionStyle{ruled}{labelfont=normalfont,labelsep=colon,strut=off} 
\floatstyle{ruled}
\newfloat{listing}{tb}{lst}{}
\floatname{listing}{Listing}
\title{AUVIC: Adversarial Unlearning of Visual Concepts for Multi-modal Large Language Models}
\author{
Haokun Chen\textsuperscript{\rm 1,2 * },  
Jianing Li\textsuperscript{\rm 3
\footnote{Equal Contribution}},
Yao Zhang\textsuperscript{\rm 1,2}, 
Jinhe Bi\textsuperscript{\rm 1},\\
Yan Xia\textsuperscript{\rm 4}\thanks{Corresponding Author},
Jindong Gu\textsuperscript{\rm 5}, 
\textbf{Volker Tresp}\textsuperscript{\rm 1,2}
}
\affiliations{
 \textsuperscript{\rm 1}LMU Munich, Munich, Germany \\
 \textsuperscript{\rm 2}Munich Center for Machine Learning (MCML), Munich, Germany \\
 \textsuperscript{\rm 3}Technical University of Munich, Munich, Germany \\
  \textsuperscript{\rm 4}University of Science and Technology of China (USTC), Hefei, China \\
 \textsuperscript{\rm 5}University of Oxford, Oxford, England \\
 chenhaokun24549@gmail.com, jianing-l@outlook.com, yan.xia@ustc.edu.cn
}

\usepackage{bibentry}

\begin{document}

\maketitle

\begin{abstract}
Multimodal Large Language Models (MLLMs) achieve impressive performance once optimized on massive datasets. Such datasets often contain sensitive or copyrighted content, raising significant data privacy concerns. Regulatory frameworks mandating the 'right to be forgotten' drive the need for machine unlearning. This technique allows for the removal of target data without resource-consuming retraining. However, while well-studied for text, visual concept unlearning in MLLMs remains underexplored. A primary challenge is precisely removing a target visual concept without disrupting model performance on related entities. To address this, we introduce AUVIC, a novel visual concept unlearning framework for MLLMs. AUVIC applies adversarial perturbations to enable precise forgetting. This approach effectively isolates the target concept while avoiding unintended effects on similar entities. To evaluate our method, we construct VCUBench. It is the first benchmark designed to assess visual concept unlearning in group contexts. Experimental results demonstrate that AUVIC achieves state-of-the-art target forgetting rates while incurs minimal performance degradation on non-target concepts.

\end{abstract}

\section{Introduction}
\begin{figure}[h!]
\centering
\hspace*{-0.015\textwidth}
\includegraphics[width=0.5\textwidth]{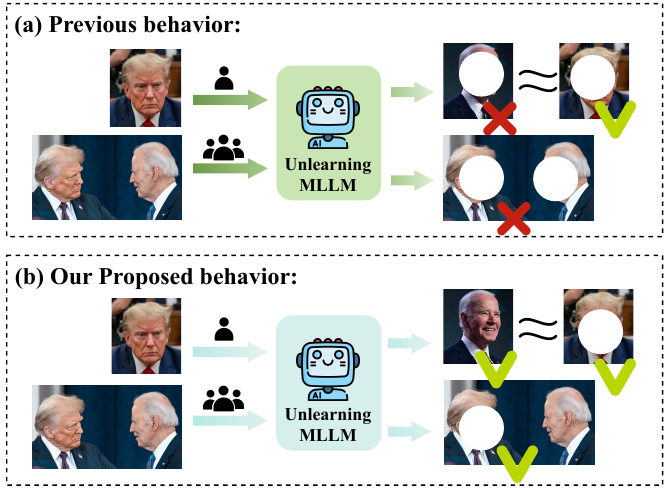}
\caption{\textbf{Comparison of unlearning behaviors in MLLMs.} (a) Existing methods fail to forget the target identity precisely in multi-person scenarios and often erase similar concepts. (b) Our method forgets the target while preserving non-target individuals, even in group settings.}

\label{fig:compare}
\end{figure}

In recent years, Multimodal Large Language Models (MLLMs) have achieved remarkable advancements, attracting significant attention from the research community. These models have excelled in a wide range of tasks, such as visual question answering \cite{kuang2025natural}. However,  their capabilities often rely on large-scale architectures, which demand extensive training on massive datasets. These datasets are typically sourced from publicly available internet corpora, which may unintentionally include harmful or inappropriate content \cite{wei2023jailbroken}. The inclusion of such data introduces potential privacy risks and the misuse of sensitive information. Moreover, regulatory frameworks like the General Data Protection Regulation (GDPR) impose legal obligations, such as the "right to be forgotten," allowing individuals to request the removal of their personal data from these models. Consequently, there is a critical need for methods that ensure these models comply with ethical principles and legal requirements.

Machine Unlearning (MU) presents a promising solution to address the challenges outlined above by selectively removing target knowledge without full retraining, it is also suitable for large-scale MLLMs. While numerous studies have explored unlearning at the granularity of entities or classes in LLMs \cite{liu2025rethinking}, precise visual concepts forgetting remains unexplored for MLLMs. A naïve approach to concept removal such as fine-tuning to suppress the target representation \cite{maini2024tofu} can lead to undesirable side effects, including degradation of overall model performance or collateral forgetting, where semantically or visually adjacent concepts are also erased.

Adapting existing MU techniques to precise concept-level forgetting in MLLMs introduces three key challenges: \textit{(C1) Utility Degradation}: Existing MU methods, e.g., Gradient Ascent, often lead to performance drops in LLMs due to poorly defined optimization objectives \cite{chen2025does}. For MLLMs, this issue is exacerbated by the complexity of aligning visual and textual modalities. Disrupting one modality can destabilize the entire generation process, resulting in hallucinated or incoherent outputs. \textit{(C2) Lack of Unlearning Data}: Effective MU relies on representative training examples. \cite{seyitouglu2024extracting} has shown that even after unlearning, residual knowledge often persists. For MLLMs, the challenge is amplified, as constructing suitable multimodal pairs for unlearning is significantly more demanding than text-only data. \textit{(C3) Collateral Forgetting}: Gradient-based MU methods often over-generalize, erasing features that are semantically or visually adjacent to the target concept. For instance, in group photographs, forgetting one individual may inadvertently erase features of surrounding individuals who share similar attributes, e.g., hairstyle, clothing, pose. These limitations are illustrated in Figure~\ref{fig:compare}, which contrasts the imprecise forgetting patterns of prior methods—often removing the target along with nearby identities in group scenes—with our approach that isolates and forgets only the designated concept while preserving others.

To address these challenges, we propose a novel method called Adversarial Unlearning of Visual Concepts (AUVIC). Specifically, we introduce a lightweight adversarial generator that perturbs input images to maximally activate the target concept, thereby providing strong supervision signals for unlearning. Furthermore, our framework leverages multimodal adversarial perturbations, including prompt variations in the text domain and feature-based image perturbations, to create challenging training data that enhance unlearning efficacy. Finally, our dynamic anchor preservation mechanism adaptively selects semantically or visually similar concepts to protect during training, effectively mitigating collateral forgetting.

Given the lack of standardized evaluation protocols for machine unlearning in MLLMs, we introduce \textit{VCUBench}, a curated benchmark specifically designed to assess fine-grained visual concept unlearning. VCUBench focuses on widely recognized public figures, containing both individual portraits and group photos. This composition allows VCUBench to robustly evaluate unlearning performance across both isolated and complex multi-person scenarios, while also testing for collateral forgetting and generalization.

Our contributions can be summarized as follows:
\begin{itemize}
    \item We propose AUVIC, a novel adversarial unlearning framework that enables precise visual concept forgetting in MLLMs by combining multimodal perturbations, feature-guided image manipulation, and a dynamic anchor preservation mechanism.

    \item We introduce VCUBench, the first benchmark for concept level unlearning in MLLMs, designed to evaluate forgetting performance and side effects in both single-person and group settings.
\end{itemize}

\section{Related Works}
\textbf{Machine Unlearning:}
Machine unlearning has emerged as a critical area of research in the context of large language models (LLMs), with various approaches proposed to address different unlearning challenges. For instance, \citet{eldan2023s} removed Harry Potter-related knowledge by fine-tuning LLMs on modified corpora with replaced keywords. \citet{zhang2024negative} optimized model preferences in the negative direction, while \citet{wang2024rkld} employed reversed knowledge distillation to erase personal information. \citet{feng2024fine} introduced a reweighted gradient ascent method, and \citet{pawelczyk2023context} demonstrated unlearning through in-context examples. Further, \citet{liu2024large, bhaila2024soft, chen2025soft} adapted input embeddings linked to unlearning targets, whereas \citet{li2024wmdp, tamirisa2024toward, huu2024effects, ashuach2024revs} proposed direct interventions in the model’s activation space. From a safety perspective, \citet{zhang2024safe, yao2023large, liu2024towards} explored methods for unlearning harmful or unsafe responses, and \citet{liang2024unlearning, liu2024efficient} focused on removing embedded backdoors. In the multimodal setting, \citet{dontsov2024clear} proposed a synthetic benchmark for evaluating unlearning in multimodal models. In this work, we propose a novel algorithm to tackle the underexplored problem of precise visual concept unlearning in multimodal large models (MLLMs).
 
\textbf{Adversarial Training:}
Several studies have investigated the use of adversarial training to enhance machine unlearning. \citet{zhang2024defensive} presented a robust unlearning framework aimed at improving diffusion model safety, while \citet{jung2024attack} introduced adversarial perturbations on unlearning samples to identify and reinitialize influential parameters. \citet{di2024adversarial} formulated a game-theoretic framework incorporating membership inference attacks (MIAs) into the unlearning process. \citet{ebrahimpour2025amun} leveraged adversarial examples to reduce model confidence in predictions on target data, enabling more accurate and safe forgetting. Additionally, \citet{peng2025adversarial} proposed mixing adversarial and clean samples to maintain model utility, and \citet{zuo2025machine} developed fine-grained perturbation strategies for selective forgetting. However, existing adversarial unlearning approaches primarily focus on textual modalities. To the best of our knowledge, our proposed method is the first to effectively address adversarially-driven multimodal unlearning, targeting fine-grained visual concepts.

\section{Motivation Analysis}
To quantify the collateral effects of unlearning in Multimodal LLMs, we conduct a controlled study using Gradient Ascent (GA) unlearning. Starting from the LLaVA-1.5 model, we apply GA on the loss of cross-entropy to a designated target concept (e.g., Donald Trump), thus reducing the confidence of the model in recognizing that concept.

We evaluate the resulting unlearned models on a dataset of \(N = 8\) concepts, each with \(n = 30\) aligned face images. For each concept \(c_i\), we independently perform GA unlearning and assess the impact of the resulting model on all concepts \(c_j\) in the evaluation set. Specifically, we define the \textit{forgetting rate} for concept \(c_j\) under a unlearned model on \(c_i\) as:
\begin{equation}
\mathcal{F}(c_j) = \frac{1}{n} \sum_{k=1}^{n} \mathbb{I} \bigl[ \hat{y}_k^{(j)} \neq c_j \bigr],
\end{equation}
where \(\hat{y}_k^{(j)}\) denotes the model’s prediction on the \(k\)-th instance of concept \(c_j\).

\begin{figure}[h!]
\centering
\includegraphics[width=0.5\textwidth]{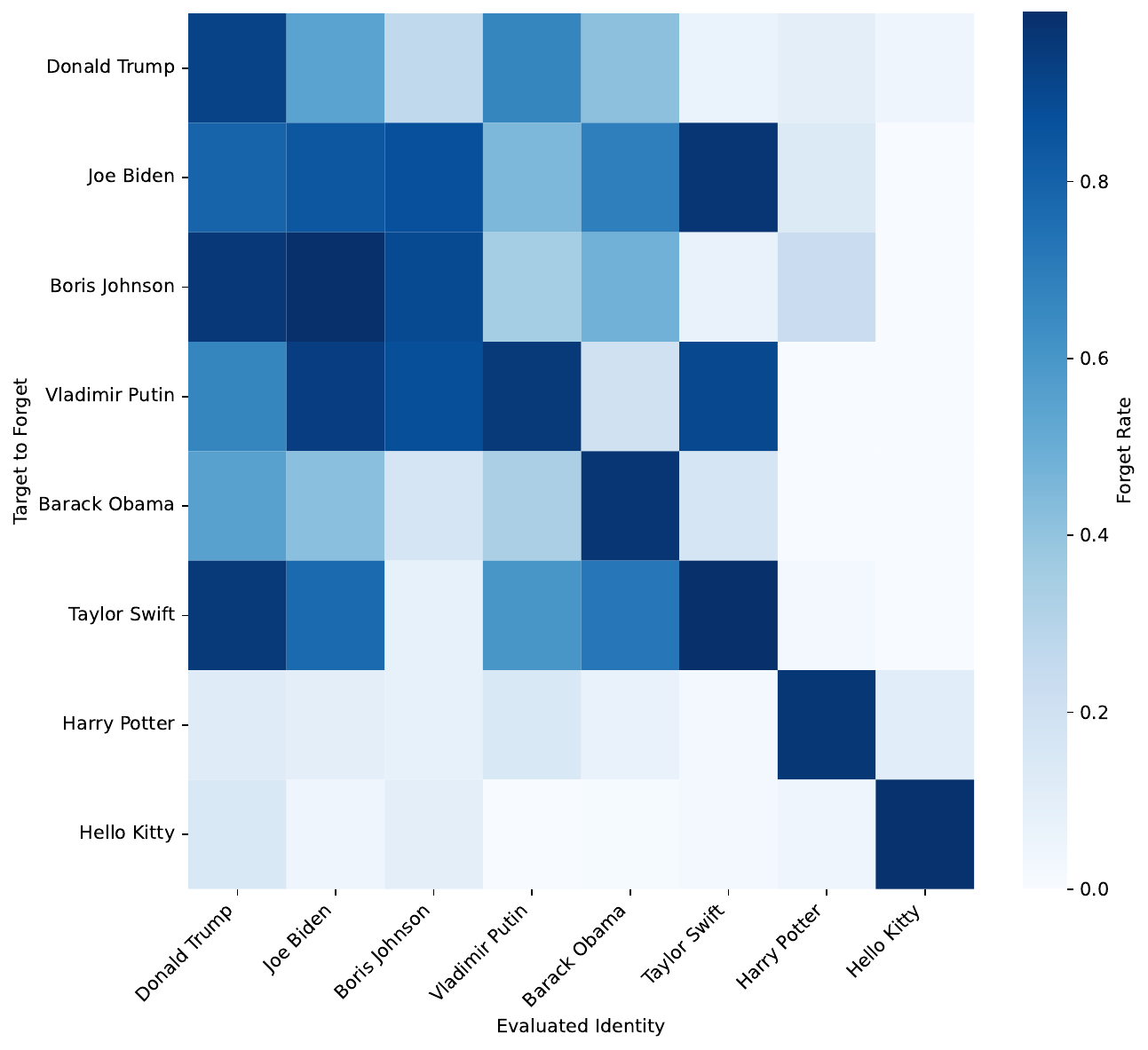}
\caption{
Collateral forgetting matrix across 8 concepts. Each row denotes an unlearned model for a specific target; each column shows unintended forgetting of other concepts.
}
\label{fig:heatmap}
\end{figure}

Figure~\ref{fig:heatmap} visualizes the forgetting rates computed from the equation above. The matrix quantifies how unlearning one concept (rows) unintentionally suppresses others (columns). We observe that visually similar concepts to the forgotten target are more susceptible to collateral degradation, revealing the overgeneralization behavior of naive gradient-ascent unlearning.

\begin{figure}[h!]
\centering
\includegraphics[width=0.4\textwidth]{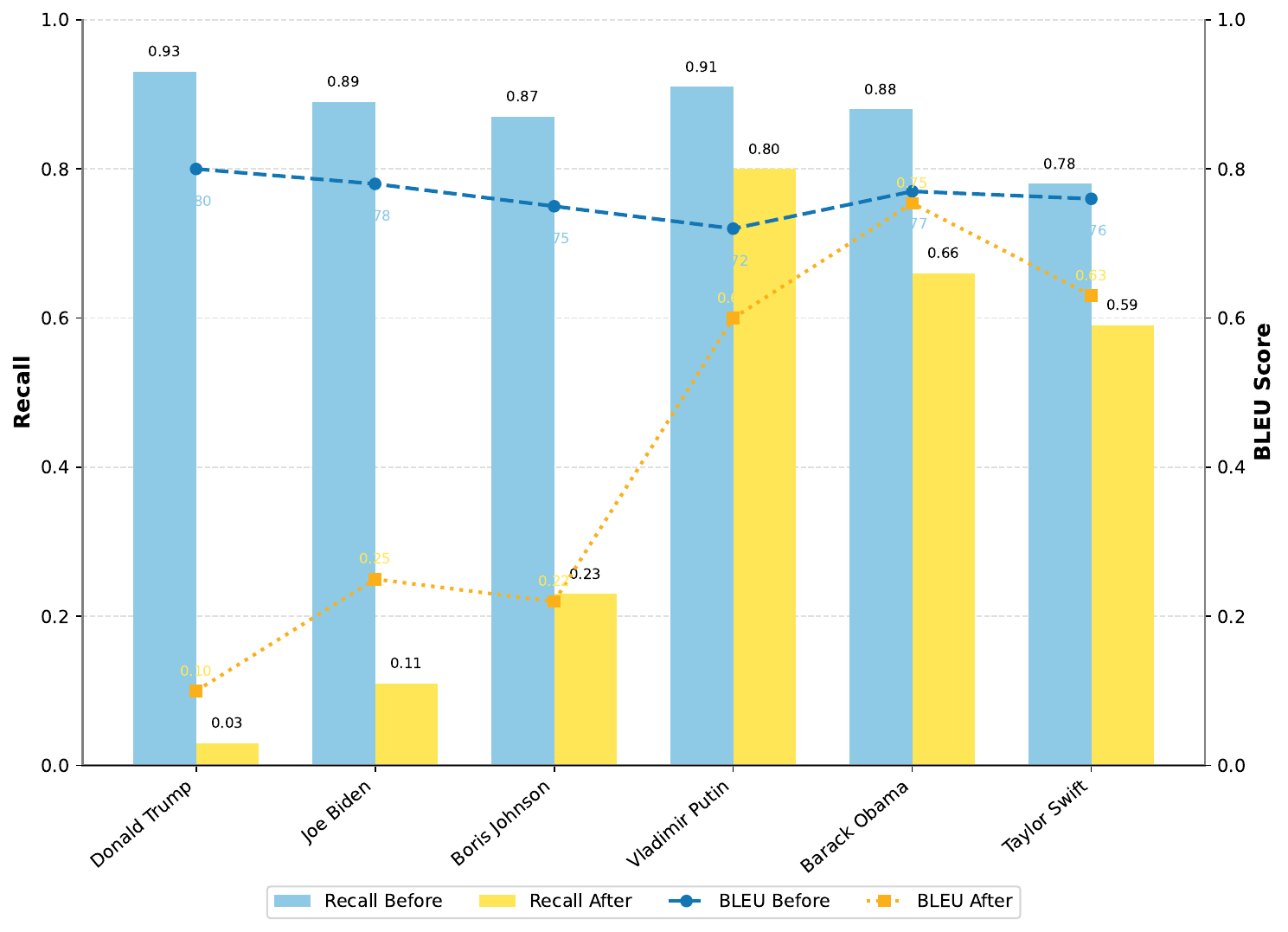}
\caption{Recall and BLEU score of each non-target concept before and after unlearning "Donald Trump"}.
\label{fig:recall}
\end{figure}
To further investigate how unlearning affects both recognition and generation, we evaluate two complementary metrics: \emph{Recall} and \emph{BLEU score} of generated captions. While recall captures the model’s ability to retain concept classification, BLEU reflects fluency and semantic alignment in textual outputs.

Figure~\ref{fig:heatmap} and Figure~\ref{fig:recall} jointly highlight the asymmetric and overgeneralized effects of GA-based unlearning. In the forgetting matrix (Figure~\ref{fig:heatmap}), we observe that removing a single concept often leads to suppression of semantically or visually similar concepts. For instance, unlearning "Donald Trump" results in elevated forgetting rates for "Joe Biden" and "Boris Johnson," indicating shared representation overlap. In contrast, visually dissimilar concepts such as "Hello Kitty" are largely unaffected, underscoring that forgetting spillover is feature-dependent rather than uniformly distributed.

These trends are further supported by the Recall and BLEU metrics in Figure~\ref{fig:recall}. The target identity "Donald Trump" is almost entirely erased (Recall: 93\% → 3\%; BLEU: 0.80 → 0.10), demonstrating the effectiveness of GA in suppressing the target. However, Recall scores for similar individuals like Biden and Johnson also drop sharply to 11\% and 23\%, respectively—revealing substantial collateral damage. In contrast, dissimilar concepts (e.g., Taylor Swift) maintain high recall and fluency, confirming the asymmetric nature of unlearning side effects.

Together, these findings show that naive unlearning strategies erase not only the target concept but also nearby features in the embedding space, leading to confusion, misclassification, and degraded caption quality. This is especially problematic in multi-concept settings such as group photographs or multi-concept dialogues, where fine-grained per-concept precision is critical.

\section{Methodology}

\begin{figure*}[ht]
\centering
\includegraphics[width=1\textwidth]{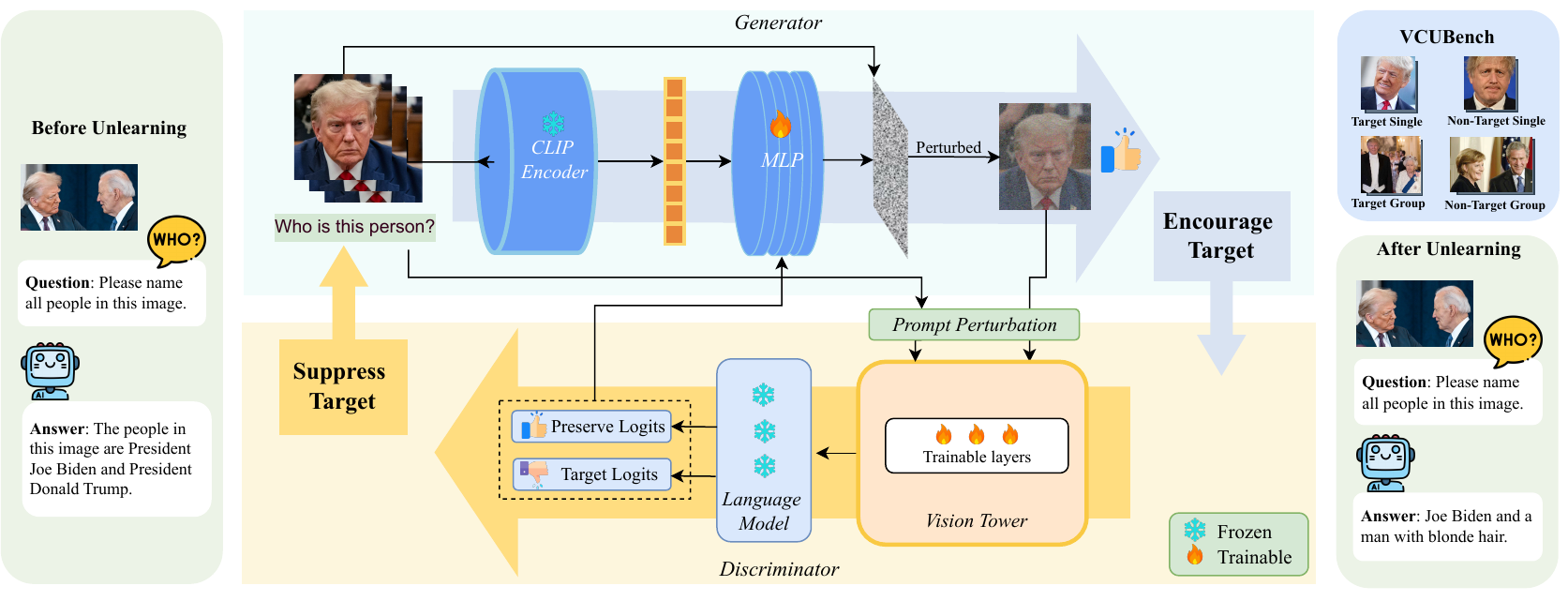}
  \caption{
  \textbf{Overview of our adversarial unlearning framework.}
  The architecture consists of a generator (top) and a discriminator (bottom) trained in an alternating loop. Given input images, the generator (top) uses a frozen CLIP encoder to extract visual features and generates perturbation through a trainable MLP. The discriminator (bottom) receives the adversarial image and a perturbed prompt, and is optimized to suppress target predictions while preserving non-target concepts. The vision tower is partially trainable, while the rest of the model remains frozen. Generator and discriminator are updated alternately in an adversarial loop. Our benchmark \textbf{VCUBench}, shown on the right, contains both single-person and multi-person images to evaluate forgetting and retention.
  }
  
\label{fig:compare}
\end{figure*}

To address this challenge, we introduce the \textbf{AUVIC} framework, which enables precise, per-concept forgetting while preserving the recognition of non-target concepts. By combining structured adversarial optimization, multimodal perturbations, and dynamic anchor preservation, AUVIC localizes erasure within the target representation without degrading the model’s overall recognition capabilities in multi-person scenarios.

\subsection{Framework Formulation}
We formulate targeted unlearning as a structured Min--Max optimization problem under adversarial perturbations. The core objective is to effectively remove the target concept while preserving the recognition of semantically related entities and maintaining general model fluency.

Specifically, we express the adversarial unlearning process as:
\begin{equation}
\min_{\theta} \max_{\phi} \; \mathbb{E}_{x \sim \mathcal{D}} 
\Big[ \mathcal{L}_{\text{f}} + \lambda \mathcal{L}_{\text{p}} + \beta \mathcal{L}_{\text{c}} \Big]
\label{eq:minmax_obj}
\end{equation}

\noindent where the loss terms are defined as:
\begin{align}
\mathcal{L}_{\text{f}} &=
\operatorname{BCE} \Bigl(
\max_{i \in \mathcal{V}(y_t)} z_\theta^{i}\bigl(G(x;\phi)\bigr),
\; \ell^{-} \Bigr),
\label{eq:loss_forget}
\\[6pt]
\mathcal{L}_{\text{p}} &=
\sum_{y_c \in \mathcal{C}} g(y_c; \tau) \cdot
\operatorname{BCE} \Bigl(
\max_{i \in \mathcal{V}(y_c)} z_\theta^{i}\bigl(G(x;\phi)\bigr),
\; \ell^{+} \Bigr),
\label{eq:loss_preserve}
\\[6pt]
\mathcal{L}_{\text{c}} &=
\operatorname{KL} \Bigl(
p_{\theta}(\cdot \mid G(x;\phi))
\;\Big\|\;
p_{\theta}(\cdot \mid x)
\Bigr),
\label{eq:loss_consistency}
\end{align}

Here, \( G(x; \phi) \) denotes the adversarial image generated by a parameterized generator \(G: \mathbb{R}^{H \times W \times 3} \rightarrow \mathbb{R}^{H \times W \times 3}\), and \( p_{\theta}(\cdot \mid \cdot) \) represents the output distribution of the vision language model with parameters \(\theta\). The target class to be forgotten is \(y_t\), and \(\mathcal{C} \subseteq \mathcal{Y} \setminus \{y_t\}\) denotes a set of semantically similar concepts that should be preserved during unlearning.

The first term, \(\mathcal{L}_{\text{f}}\), is the \textit{target forgetting loss}, which computes the binary cross-entropy (BCE) between the model's maximal prediction logits over the visual tokens corresponding to \(y_t\) and a suppression target \(\ell^{-}\). It is maximized by the generator to strongly activate the forgotten concept and minimized by the model to suppress it, thereby enabling adversarial forgetting.

The second term, \(\mathcal{L}_{\text{p}}\), is the \textit{concept preservation loss}, which encourages the model to retain accurate predictions on protected concepts \(y_c \in \mathcal{C}\). Each class is weighted by a Gumbel-Softmax factor \(g(y_c; \tau)\), which provides a smooth and differentiable approximation of top-\(k\) sampling. This weighting ensures that preservation is focused on the most semantically proximate concepts, and the BCE between predicted logits and a positive target \(\ell^{+}\) is minimized to maintain recognition.

The third term, \(\mathcal{L}_{\text{c}}\), is a \textit{consistency regularization loss}, defined as the KL divergence between the model's predictive distributions on clean versus adversarial inputs. This term penalizes unnecessary distributional shifts, encouraging stable generation and fluency in downstream tasks.

This formulation enables both the generator and the discriminator to be updated via gradient-based methods, where the generator persistently produces perturbations in the input space to activate the target concept, while the discriminator learns to suppress target concept activation under adversarial conditions without degrading performance on semantically close concepts. By leveraging the Gumbel-Softmax mechanism, we efficiently select relevant concepts for protection in a differentiable manner, allowing our adversarial unlearning framework to achieve targeted forgetting with high specificity and minimal collateral forgetting.

\subsection{Multimodal Perturbations}
To facilitate effective unlearning of a specific visual concept, we first augment the training inputs using adversarial perturbations in both modalities.

On the textual side, we employ prompt perturbation to enhance adversarial robustness during optimization. We generate diverse, semantically equivalent queries by introducing controlled variations, such as greetings, rephrasings, and contextual distractors. Moreover, we prepend the target concept’s name to each query to ensure that the model's attention is explicitly anchored to the concept to be forgotten. This explicit grounding directs the model’s attention to the target concept, amplifying its activation and providing stronger gradients during adversarial optimization.

On the visual side, we introduce a \emph{feature-guided image perturbation} module to generate adversarial examples that emphasize the target concept. Specifically, we freeze a pretrained CLIP image encoder and extract a visual feature vector \( h = \mathrm{CLIP}(x) \) from each clean image \(x\).

A lightweight generator network \( G_\theta \), comprising three \texttt{Linear} layers interleaved with \texttt{ReLU} activations and a final \(\tanh\) nonlinearity, maps the feature vector \(h\) to an image-shaped perturbation \( \delta_\theta \in \mathbb{R}^{3 \times H \times W} \). The output is passed through a \(\tanh\) activation and scaled to control the perturbation magnitude, yielding the final adversarial image:
\begin{equation}
x' = \operatorname{clip}\left(x + \delta_\theta,\, 0, 1\right),
\end{equation}
where pixel values are clipped to remain within valid image bounds. This construction ensures that the perturbation remains visually imperceptible by enforcing an \(\ell_\infty\) constraint on \( \delta_\theta \).

\subsection{Parameter-Efficient Dynamic Anchor Preservation}

We perform adversarial unlearning in a \emph{parameter-efficient} manner by updating only the vision subsystem while preserving linguistic fluency. Specifically, we insert LoRA adapters~\citep{hu2021loralowrankadaptationlarge} into the CLIP vision tower of LLaVA-1.5 while keeping all original weights frozen. For each weight matrix \(W_{0} \in \mathbb{R}^{d_{\text{out}} \times d_{\text{in}}}\), we learn a low-rank decomposition with rank \(r=32\) and scaling factor \(\alpha=32\). 

To mitigate collateral forgetting of semantically related concepts during targeted unlearning, we introduce a \emph{dynamic anchor preservation} mechanism based on Gumbel-Softmax sampling. Given a GPT-generated list of \(K\) public-figure names, we compute their mean token embeddings \(\{\bar{e}_i\}_{i=1}^K\) from the frozen vocabulary table \(\mathbf{E} \in \mathbb{R}^{V \times d}\). Let \(\bar{e}_T\) denote the mean embedding of the target concept. We then compute cosine similarities between the target and each candidate \( s_i \)
, then add Gumbel noise \(g_i\), and apply a temperature-controlled softmax to yield a sampling distribution. The final preserve set \(\mathcal{P}_{\text{top}}\) is defined as:
\begin{equation}
\mathcal{P}_{\text{top}} = \mathrm{Top}_m\left( 
\mathrm{GumbelSoftmax}(\{s_i + g_i\}; \tau) 
\right),
\end{equation}
where \(\tau\) controls the sharpness of the distribution. This mechanism adaptively identifies \(m\) semantically related concepts to be protected during unlearning.

\section{Benchmark}
\subsection{Dataset}
\label{sec:dataset}

VCUBench is a carefully constructed benchmark that targets public single and group figures. For each concept, VCUBench collects single photographs of himself/herself (\textit{Target-Single}),  single photographs of other celebrities as a control (\textit{Non-Target-Single}), and group photographs of multiple people (\textit{Target-Group}) containing the target; and uniformly provides group scenes without the five targets at all (\textit{Non-Target-Group}) for negative-sample evaluation.

To guarantee label fidelity, every picture is verified by the original LLaVA-1.5 to filter out the model misidentifies.  We further augment each retained image with twenty VQA-style queries that probe presence, concept, spatial relation, and commonsense context.  The resulting release contains 15k+ high-quality image–question–answer triples and serves as the first dedicated benchmark for evaluating targeted visual-concept unlearning in multimodal LLMs.

\subsection{Metrics}
\label{sec:metrics}
To evaluate AUVIC and all baselines on \textbf{VCUBench}, we track
three complementary axes: targeted forgetting performance, non-target retention, and language fluency.
The quantitative indicators listed in
Table~\ref{tab:unlearning_results} are defined as follows:

\subsubsection {Target Forgetting Accuracy (TFA).}
This metric measures whether the model successfully suppresses recognition of the target concept in multi-person settings. It is defined as 
\(\text{TFA} = N_{\text{non-} c_t} / N_{c_t}\), 
where \(N_{c_t}\) is the total number of group images that contain the target person, and \(N_{\text{non-} c_t}\) is the number of such images where the model fails to mention or identify the target. A higher TFA indicates stronger forgetting.

\subsubsection{Non-Target Retain Accuracy (NTRA).}
This dimension measures the extent to which the model preserves recognition of non-target concepts after unlearning is applied to a specific target. Formally, NTRA is defined as \(N_{\text{correct-}c_{\neg t}} / N_{c_{\neg t}}\), where \(N_{c_{\neg t}}\) denotes the number of non-target instances, and \(N_{\text{correct-}c_{\neg t}}\) is the number of those instances where the model correctly identifies the non-target individual. A higher NTRA indicates stronger preservation of non-target knowledge.

\subsubsection {Group Retain–Forget F\textsubscript{1}.}
A harmonic mean balancing TFA and NTRA:
\begin{equation}
\text{GRF-F1} =
\frac{2 \times \text{TFA} \times \text{NTRA}}{\text{TFA} + \text{NTRA}}.
\label{eq:GRF}
\end{equation}
This rewards methods that achieve high forgetting with minimal collateral damage.

\subsubsection{Efficacy.}
This metric measures unlearning effectiveness using single person images of the target identity. For each input, we compute an exact match (EM) between the model's response and the ground-truth label. Efficacy is defined as the proportion of samples where the model fails to identify the target:\begin{equation}
\text{Efficacy} = \frac{N_{\text{forgotten}}}{N_{\text{total}}},
\label{eq:efficacy}
\end{equation}
where \(N_{\text{forgotten}}\) denotes the number of incorrect or abstained predictions. Higher values indicate stronger forgetting.

\paragraph{Generality.}
To assess whether unlearning impacts general visual reasoning, we evaluate models on a held-out split from ScienceVQA~\cite{lu2022learnexplainmultimodalreasoning}. Higher accuracy reflects stronger generalization and minimal side effects on unrelated areas.

\begin{table*}[ht]
    \centering
    \renewcommand{\arraystretch}{1.2}
    \setlength{\tabcolsep}{6pt} 
    \caption{Quantitative results on the targeted unlearning benchmark with \textit{Trump} as the example target concept. Higher TFA, NTRA, and GRF-F1 indicate better targeted forgetting and non-target preservation.}
    \label{tab:unlearning_results}
    \small
    \begin{tabularx}{0.9\textwidth}{ccccccc}
        \toprule
        \textbf{Method} & \textbf{TFA (\%, $\uparrow$)} & \textbf{NTRA (\%, $\uparrow$)} & \textbf{GRF-F1 (\%, $\uparrow$)} & \textbf{Efficacy ($\uparrow$)} & \textbf{Generality ($\uparrow$)} & \textbf{Perplexity ($\downarrow$)} \\
        \midrule
        GA & 84.48\% & 30.17\% & 44.46\% & 89.17\% & 63.07 & 16.39 \\
        PO & 49.14\% & 54.48\% & 51.67\% & 80.42\% & 62.91 & 7.5779 \\
        GA+KL & 85.86\% & 26.55\% & 40.56\% & 90.62\% & 62.98 & 8.92 \\
        SIU & 92.35\% & 63.49\% &75.25\% & 100.0\% & 61.2\% & 11.26 \\
        \textbf{AUVIC (Ours)} & \textbf{93.64\%} & \textbf{83.17\%} & \textbf{88.10\%} & \textbf{97.92\%} & \textbf{63.05\%} & \textbf{8.1413} \\
        \bottomrule
    \end{tabularx}
\end{table*}

\paragraph{Perplexity.}
To evaluate whether unlearning affects the model's language fluency, we compute the perplexity of generated captions:
\begin{equation}
\text{PPL} = \exp\left(-\frac{1}{T} \sum_{t=1}^T \log P_{\hat{\mathcal{M}}}(w_t \mid w_{<t}, I) \right),
\label{eq:ppl}
\end{equation}
where \(I\) is the input image and \(w_t\) is the \(t\)-th token in the caption.
To avoid penalizing the model for omitting deliberately forgotten names, we mask all tokens \(w_t \in \mathcal{C}\) when computing perplexity:
\begin{equation}
P_{\hat{\mathcal{M}}}^{\text{mask}}(w_t \mid \cdot) = 
\begin{cases}
P_{\hat{\mathcal{M}}}(w_t \mid \cdot), & w_t \notin \mathcal{C} \\
\frac{1}{V}. & w_t \in \mathcal{C}
\end{cases}
\label{eq:ppl_mask}
\end{equation}

\section{Experiments}

\subsection{Experiment Setup}

We evaluate \textbf{AUVIC} and baselines on \textbf{VCUBench}, a benchmark designed for fine grained unlearning of visual concepts. It includes five public figures across both single-person and group-photo settings to support detailed diagnosis of forgetting and retention.

All models are initialized from LLaVA-1.5 (7B). We apply LoRA adapters (\(r = 32\), \(\alpha = 32\)) to the vision encoder, specifically targeting the \texttt{q\_proj}, \texttt{v\_proj}, \texttt{fc1}, and \texttt{fc2} modules. Training is conducted on 3× RTX 4090 GPUs. We use AdamW optimizers with ReduceLROnPlateau schedulers for both the vision encoder and the generator, optimized independently.

The adversarial generator comprises three stacked \texttt{Linear}–\texttt{ReLU} layers with hidden dimension 1024. It maps CLIP image features to perturbations bounded by \(\varepsilon = 8/255\) in \(\ell_\infty\) norm. Perturbations are applied at \(336 \times 336\) resolution and clipped to ensure valid pixel values.

\subsection{Baseline}
We evaluate AUVIC against the following representative unlearning approaches: 
(i) \textbf{Preference Optimization (PO):} Inspired by the TOFU framework~\cite{maini2024tofu}, we fine-tune the model to prefer abstention-style responses (e.g., ``I don't know.'') when prompted with questions involving the target concept \(C\). 
(ii) \textbf{Gradient Ascent (GA):} Following~\cite{liu2025rethinking}, this method performs targeted forgetting by maximizing the prediction loss on samples involving \(C\), directly suppressing the model’s confidence in the target. 
(iii) \textbf{GA with KL Regularization (GA+KL):} To mitigate degradation of general performance, this baseline augments GA with a KL-divergence penalty~\cite{zhang2024negative} between predictions on clean and perturbed data, encouraging stability outside the unlearning region.

\subsection{Experiment Results}

\begin{table*}[ht]
    \centering
    \renewcommand{\arraystretch}{1.2}
    \setlength{\tabcolsep}{6pt}  
    \caption{Quantitative results on the targeted unlearning benchmark across six target concepts.}
    \label{tab:avg_unlearning_results}
    \small
    \begin{tabularx}{0.8\textwidth}{ccccccc}
        \toprule
        \textbf{Method} & \textbf{TFA ($\uparrow$)} & \textbf{NTRA ($\uparrow$)} & \textbf{GRF-F1 ($\uparrow$)} & \textbf{Efficacy ($\uparrow$)} & \textbf{Generality ($\uparrow$)} & \textbf{Perplexity ($\downarrow$)} \\
        \midrule
        GA & 67.67 & 30.12 & 37.64 & 77.76 & 59.62 & 18.47 \\
        PO & 49.87 & 55.91 & 50.19 & 70.71 & 61.62 & 9.86 \\
        GA+KL & 77.36 & 32.99 & 43.80 & 82.95 & 60.83 & 11.06 \\
        \textbf{AUVIC (Ours)} & \textbf{96.99} & \textbf{75.34} & \textbf{84.94} & \textbf{96.82} & \textbf{62.69} & \textbf{8.34} \\
        \bottomrule
    \end{tabularx}
\end{table*}

Compared to existing unlearning methods, AUVIC achieves the best trade-off between precision forgetting and collateral preservation. As shown in Table~\ref{tab:unlearning_results}, AUVIC consistently improves both TFA and NTRA, yielding the highest Group Retain–Forget F1. This demonstrates its ability to isolate the target concept while preserving the recognition and generation capabilities of semantically similar individuals.

In particular, AUVIC achieves a TFA of 93.64 and an NTRA of 83.17, outperforming GA and GA+KL, which suffer from substantial degradation in non-target performance (e.g., GA: NTRA 30.17). While SIU achieves competitive forgetting, its retention score remains lower, indicating insufficient precision in visual subspace isolation. Furthermore, AUVIC maintains performance on general VQA tasks (Generality) and achieves the lowest Perplexity, confirming that our visual-side fine-tuning strategy does not compromise the linguistic generation capabilities of the MLLM.

Table~\ref{tab:avg_unlearning_results} presents the average performance of different unlearning methods evaluated across six target concepts. We observe that our proposed method, AUVIC, achieves the best overall performance across all six metrics. It obtains the highest TFA, NTRA, GRF-F1, and Efficacy, indicating a superior ability to erase targeted knowledge while preserving the generalization capacity and functionality on non-target concepts. Furthermore, AUVIC maintains a low perplexity score, suggesting minimal degradation in language fluency and generation quality.

\subsection{Example Outputs}
\begin{figure}[htb]        
  \centering
  \includegraphics[width=0.35\textwidth]{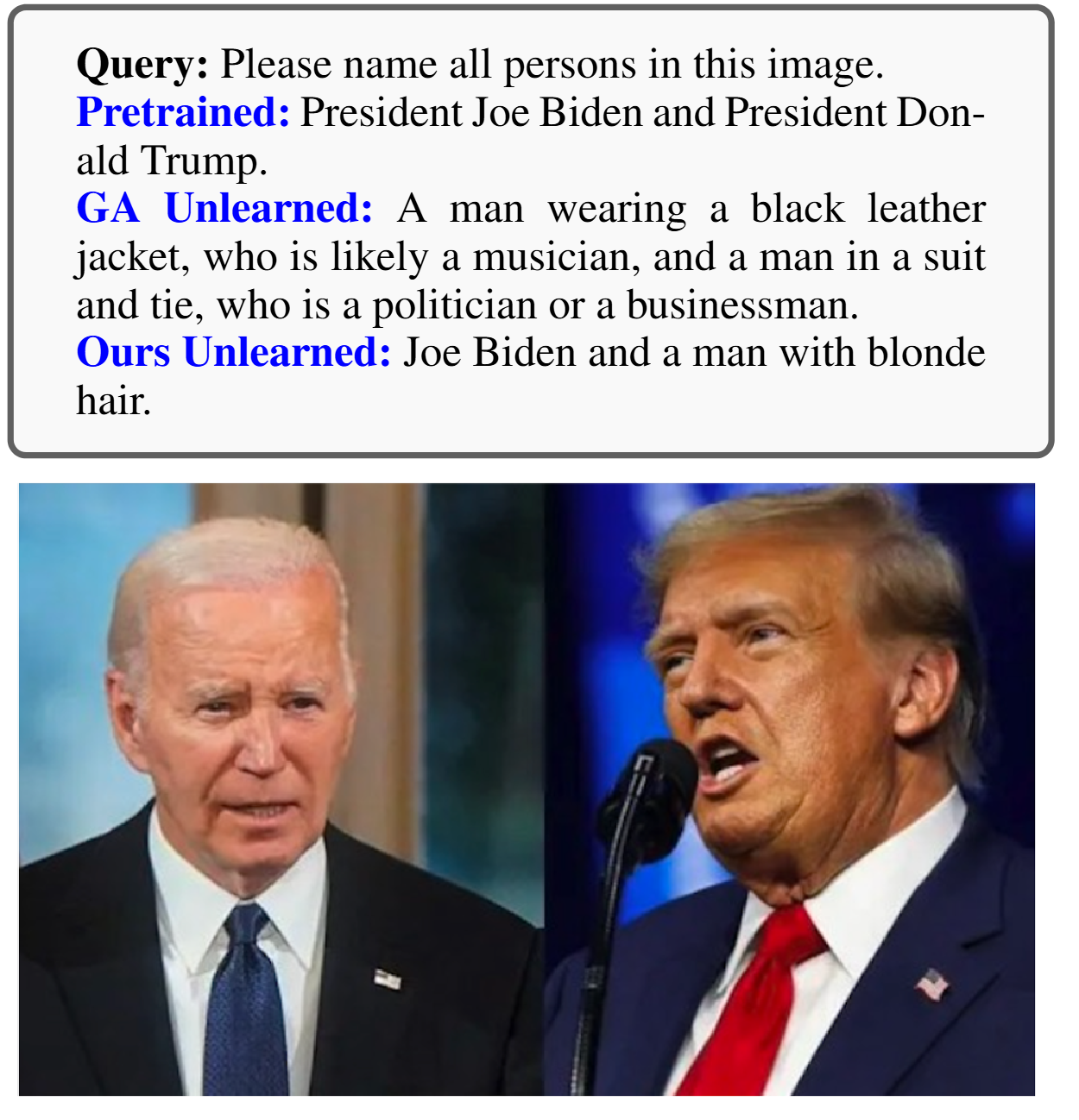}  
  \caption{Example evaluation responses of both GA and proposed method in our benchmark. Unlearning Target: \textit{Donald Trump}.}
  \label{fig:example}
\end{figure}

In this section, we compare the output between the proposed method and the baseline. As shown in Figure~\ref{fig:example}, we observe that the pretrained model correctly names both individuals. After GA unlearning, the model forgets both individuals and offers vague descriptions. The proposed unlearning method more precisely removes Trump's identity, referring to him only as “a man with blonde hair,” while still correctly naming Joe Biden. This observation further indicates the effectiveness of the proposed method in precise unlearning.

\subsection{Ablation Study}
We perform a systematic ablation to evaluate the significance of two core mechanisms in AUVIC: (1) the use of adversarial perturbations to drive forgetting and (2) the use of Gumbel-Softmax sampling for dynamic anchor preservation. Each mechanism is removed independently to assess its effect on forgetting specificity and retention capability, as summarized in Table~\ref{tab:ablation}. We observe that adding the adversarial perturbation leads to the most performance gain.

\begin{table}[b]
\centering
\small
\caption{Ablation results on VCUBench (\textit{Trump} as the target). Gumbel sampling and adversarial perturbation both contribute to effective and precise forgetting.}
\label{tab:ablation}
\begin{tabularx}{0.95\linewidth}{X ccc}
\toprule
\textbf{Variant} & \textbf{TFA} & \textbf{NTRA} & \textbf{GRF-F1}  \\
\midrule
w/o Gumbel & 89.14 & 64.57 & 72.37 \\
w/o Adv Perturb & 83.2 & 60.43  & 70.98 \\
w/o Both & 27.43 & 75.43 & 38.55 \\
\bottomrule
\end{tabularx}
\end{table}

\section{Conclusion}
In this work, we introduced AUVIC, a novel framework designed to address the critical challenge of precisely unlearning visual concepts from Multimodal Large Language Models (MLLMs). By employing adversarial perturbations and a dynamic anchor preservation mechanism, AUVIC effectively removes targeted information while preventing the unintended erasure of related concepts—a common issue known as collateral forgetting. To validate our approach, we developed VCUBench, the first benchmark specifically for evaluating visual concept unlearning. The experimental results confirm that AUVIC achieves a state-of-the-art balance between forgetting the target concept and retaining knowledge of non-target entities, demonstrating its ability to perform precise and isolated unlearning without degrading the model's overall performance or language fluency.

\bibliography{aaai2026}

\end{document}